\documentclass{article}

\usepackage[final]{neurips_2024}

\usepackage[utf8]{inputenc} 
\usepackage[T1]{fontenc}    
\usepackage{hyperref}       
\usepackage{url}            
\usepackage{booktabs}       
\usepackage{amsfonts}       
\usepackage{nicefrac}       
\usepackage{microtype}      
\usepackage{xcolor}         
\usepackage{graphicx}
\usepackage{enumitem}       
\usepackage{subcaption}
\usepackage{array}
\usepackage{tabularx}
\setcitestyle{square,comma,numbers,sort} 

\title{Long Context RAG Performance of Large Language Models}

\author{%
  Quinn Leng\thanks{Equal contribution} \\
  Databricks Mosaic Research\\
  \texttt{quinn.leng@databricks.com} \\
  \And
  Jacob Portes$^*$ \\
  Databricks Mosaic Research\\
  \texttt{jacob.portes@databricks.com} \\
  \AND
  Sam Havens \\
  Databricks Mosaic Research\\
  \texttt{sam.havens@databricks.com} \\
  \And
  Matei Zaharia \\
  Databricks Mosaic Research\\
  \texttt{matei@databricks.com} \\
  \And
  Michael Carbin \\
  Databricks Mosaic Research\\
  \texttt{michael.carbin@databricks.com}
}

\begin{document}

\maketitle
\begin{abstract}
Retrieval Augmented Generation (RAG) has emerged as a crucial technique for enhancing the accuracy of Large Language Models (LLMs) by incorporating external information. With the advent of LLMs that support increasingly longer context lengths, there is a growing interest in understanding how these models perform in RAG scenarios. Can these new long context models improve RAG performance? This paper presents a comprehensive study of the impact of increased context length on RAG performance across 20 popular open source and commercial LLMs. We ran RAG workflows while varying the total context length from 2,000 to 128,000 tokens (and 2 million tokens when possible) on three domain-specific datasets, and report key insights on the benefits and limitations of long context in RAG applications. Our findings reveal that while retrieving more documents can improve performance, only a handful of the most recent state of the art LLMs can maintain consistent accuracy at long context above 64k tokens. We also identify distinct failure modes in long context scenarios, suggesting areas for future research.
\end{abstract}

\setcounter{footnote}{0} 

\section{Introduction} 

The development of Large Language Models (LLMs) with increasingly longer context lengths has opened new possibilities for Retrieval Augmented Generation (RAG) applications. Recent models such as Anthropic Claude (200k tokens) \citep{anthropic2023claude}, GPT-4-turbo (128k tokens) \citep{achiam2023gpt}, OpenAI o1 (128k tokens) \citep{OpenAI2024}, Llama 3 \citep{dubey2024llama} and Google Gemini 1.5 Pro (2 million tokens) \citep{reid2024gemini} have led to speculation about whether long context models might eventually subsume traditional RAG workflows entirely. In this study, we empirically investigate the impact of increased context length on RAG performance and explore the limitations and challenges that arise in long context scenarios.

RAG can enhance the accuracy of LLMs by retrieving information from external sources, enabling users to incorporate task-specific or private data into their LLM workflows. 
Published results using RAG-like methods have demonstrated benefits across many applications \citep{kim2024retrieval} including machine translation \citep{khandelwal2021nearest}, semantic parsing \citep{Zhuo2023OnRO}, question answering \citep{asai2022task,chen2023augmenting,Lee2019LatentRF,Nakano2021WebGPTBQ}, and open-ended text generation \citep{khandelwal2019generalization}. With longer context lengths, LLM developers can feed more documents into their RAG applications. While there has been recent speculation that long context LLMs will \textit{replace} RAG entirely \citep{lee2024can}, in this paper we study whether long context LLMs can indeed be \textit{used effectively }for RAG systems. How well do the best open source and commercial models do on long-context RAG tasks?

In this study, we apply a standard RAG approach and evaluate the performance of 20 popular open source and commercial LLMs with varying context lengths from 2,000 to 128,000 tokens (and 2 million tokens when possible). We then analyze distinct failure modes for different models across long context RAG scenarios. We show that:

\begin{itemize}
    \item \textbf{Using longer context does not uniformly increase RAG performance}. The majority of models we evaluated first increase and then decrease RAG performance as context length increases. Only a handful of the most recent state of the art LLMs can maintain consistent accuracy at long context above 64k tokens.
    \item \textbf{LLMs fail at long context RAG in unique ways as a function of context length}. While some models tended to provide incorrect answers, others failed to follow instructions or refused to answer due to perceived copyright concerns.
\end{itemize}


\begin{figure}
    \centering
    \includegraphics[width=1\linewidth]{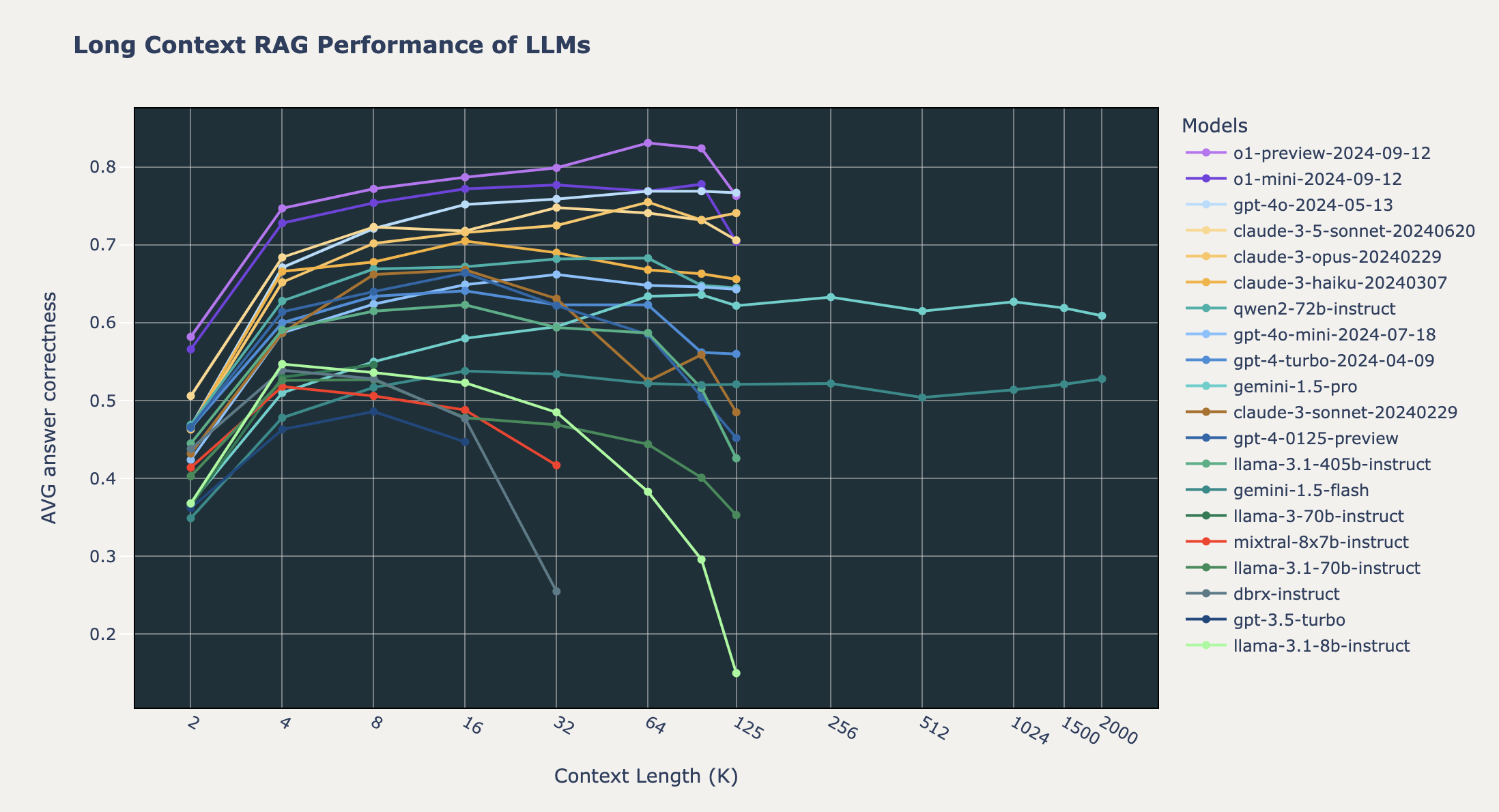}
    \caption{Long context RAG performance of o1, GPT-4, Claude 3/3.5, Gemini 1.5 (\texttt{gemini-1.5-pro-001} and \texttt{gemini-1.5-flash-001}), Llama 3/3.1, Qwen 2, Mistral and DBRX models on 3 curated RAG datasets (Databricks DocsQA, FinanceBench, and Natural Questions). All values can be found in Table \ref{appendix:results-table}. Model versions are listed in Table \ref{appendix:llm-models}.}
    \label{fig:long-context-performance}
\end{figure}

\vspace{-10pt}
\section{Background and Related Work} 

RAG combines the strengths of retrieval-based and generation-based approaches in natural language processing, and has shown significant improvements in the quality of question-answering systems across many domains and tasks \citep{lewis2020retrieval, karpukhin2020dense, gao2023retrieval}. The process typically involves two main steps, retrieval and generation. During the first stage,  relevant information is retrieved from a corpus or database based on a user query. This often involves embedding documents and storing them in a vector database for similarity-based retrieval. During the generation stage, the retrieved information is combined with the user query as input to an LLM. Importantly, multiple retrieved documents can be included as input to the LLM depending on the maximum context length of the model.

Recent advancements in LLM capabilities have led to models with increasingly larger context lengths. While early models like GPT-3.5 had a context length of 4k tokens, newer models such as Anthropic Claude (200k tokens), OpenAI o1 (128k tokens) and Google Gemini 1.5 models (2 million tokens) support much longer contexts. Open source models have followed a similar trend, with recent models like Mixtral \citep{jiang2024mixtral} and DBRX \citep{mosaic2024dbrx} supporting 32k tokens, and Llama 3.1 reaching 128k tokens.

However, recent studies have identified limitations in long context models. For example, the ``lost in the middle'' paper \citep{zhang2023lost} found that models struggle to retain and utilize information from the middle portions of long texts, leading to performance degradation as context length increases. Similarly, the RULER paper \citep{zhang2023ruler} found that the ``effective context length'' (usable context before performance decreases) can be much shorter than the claimed maximum context length. Recent studies have also tried to compare RAG to workflows where the entire corpus is included in the context window of the LLM \citep{li2024retrieval}. This has only been possible to do with the very recent state of the art models such as o1, GPT-4o, Claude 3.5, Gemini 1.5, Qwen 2 72B and Llama 3.1 405B, and the jury is still out on whether such an approach leads to accurate results and is cost effective. Other relevant studies and blogposts include \cite{sun2021long,laban2024summary, lee2024can, KirkovskaSeethepalli, AlammarVoisinBarnett, li2024retrieval, yue2024inference, goldman2024really}. Similar to our study, \citeauthor{jin2024long} find that increasing
the number of retrieved passages does not consistently improve RAG performance for Gemma-7B, Gemma-2-9B, Mistral NeMo 12B but does for Gemini 1.5 Pro \citep{jin2024long}. Our concurrent work corroborates this across 20 closed and open source models.

\vspace{-10pt}

\section{Methodology} 
\vspace{-5pt}
We conducted RAG experiments using 20 popular open source and commercial LLMs, and evaluated their performance on three datasets: Databricks DocsQA,\footnote{Databricks DocsQA is a benchmark of technical questions and answers related to the Databricks platform.} FinanceBench \citep{islam2023financebench}, and Natural Questions \citep{kwiatkowski2019natural}. For the retrieval stage, we retrieved document chunks using the same embedding model across all settings (OpenAI \texttt{text-embedding-3-large}\footnote{\url{https://openai.com/index/new-embedding-models-and-api-updates/}} with a chunk size of 512 tokens and a stride of 256 tokens) and used FAISS\footnote{\url{https://github.com/facebookresearch/faiss}} (with IndexFlatL2 index) as the vector store. These chunks were then inserted into the context window of a generative model.

We then evaluated how generation performance changes as a function of the number of retrieved document chunks by varying the LLM context from 2,000 tokens to 128,000 tokens (and 2 million tokens when possible). We evaluated the following models: o1-mini, o1-preview, Gemini 1.5 Pro, Gemini 1.5 Flash,\footnote{We used the versions of Gemini 1.5 released in June 2024, specifically \texttt{gemini-1.5-pro-001} and \texttt{gemini-1.5-flash-001} with 2 million token context windows.} GPT-4o, Claude 3.5 Sonnet,\footnote{We used the Claude 3.5 Sonnet released in June 2024, \texttt{claude-3-5-sonnet-20240620}} Claude 3 Opus, Claude 3 Sonnet, Claude 3 Haiku, GPT-4o mini, GPT-4 Turbo, GPT-4, Llama 3.1 405B, Llama 3 70B, Llama 3.1 70B, Llama 3.1 8B, Qwen 2 72B, Mixtral 8x7B, DBRX, and GPT-3.5 Turbo. These models represent some of the most popular API-based and open source LLMs as of this writing. A full list of the model versions used in this study can be found in Table \ref{appendix:llm-models}. 

For generation, we set the temperature to 0.0 and the maximum output sequence length to 1024. We used a simple prompt template to combine the retrieved documents with the user query for each model and dataset (Appendix \ref{appendix:prompts}). The system had to correctly answer questions based on the retrieved documents, and the answer was judged by a calibrated ``LLM-as-a-judge'' using GPT-4o (see Appendix \ref{appendix-llm-as-judge} for further details).

Finally, we analyzed the failure patterns for selected models (OpenAI o1, Gemini 1.5 Pro, Llama 3.1 405B, GPT-4, Claude 3 Sonnet, DBRX, and Mixtral) in long context scenarios by using GPT-4o to classify failures into broad categories such as ``refusal'' and ``wrong answer'' (Appendix \ref{appendix:prompt-failure-analysis}). We also include an analysis of retrieval performance (recall@k) in Appendix \ref{appendix-retrieval-performance}.

\begin{figure}[t!]
    \centering
    \includegraphics[width=1\linewidth]{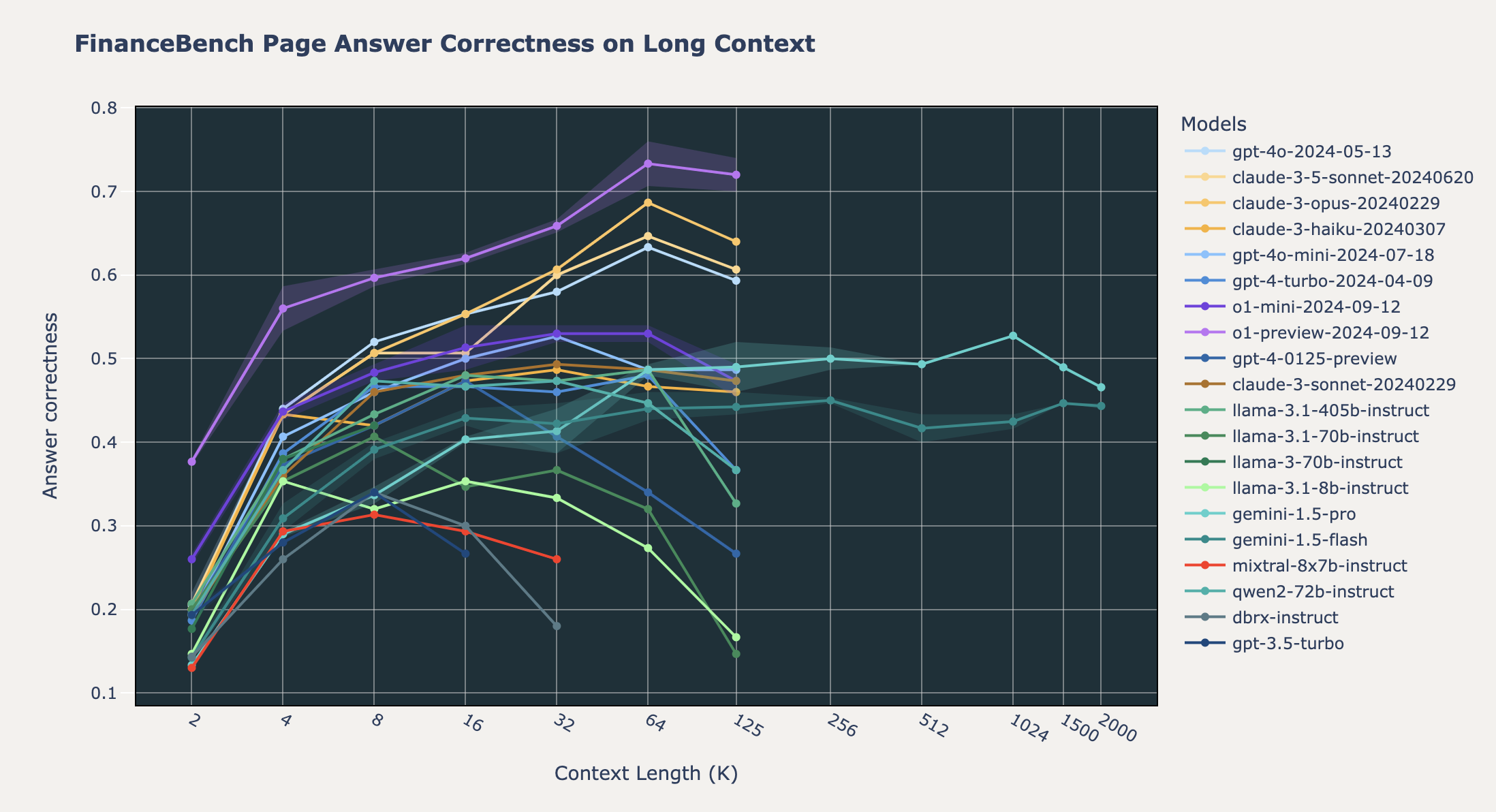}
    \caption{Long context RAG performance on FinanceBench}
    \label{fig:fb-long-context-performance}
\end{figure}

\vspace{-10pt}
\section{Results}

\subsection{Using longer context does not uniformly increase RAG performance}

The best commercial models such as o1-mini/preview, GPT-4o, and Claude 3.5 Sonnet steadily improve performance as a function of context length, while the majority of the open source models first increase and then decrease performance as context length increases (Figs. \ref{fig:long-context-performance} and \ref{fig:fb-long-context-performance}). Overall, we found that the following models show consistent accuracy improvement up to 100k tokens: o1-preview and o1-mini, GPT-4o and GPT-4o mini, Claude 3.5 Sonnet, Claude 3 Opus, and Gemini 1.5 Pro. These models exhibit largely monotonic behavior where the results don't get significantly worse after they peak.

Among the open source models, Qwen 2 70B maintains consistent accuracy up to 64k. Llama 3.1 405B performance starts to decrease after 32k tokens, GPT-4-0125-preview starts to decrease after 64k tokens, and only a few models can maintain consistent long context RAG performance on all datasets. This demonstrates that while some models that boast long contexts can be used effectively to increase RAG performance, the majority of open source models can only handle effective RAG tasks up to roughly 16k-32k tokens.

We report very strong performance from the OpenAI o1 models; the o1 models seem to be a substantive improvement over GPT-4 and GPT-4o. Although the overall answer correctness of the Google Gemini 1.5 Pro and Gemini 1.5 Flash models is much lower than that of the o1 and GPT-4o models up to 128,000 tokens, the Gemini models maintain consistent performance at extremely long contexts up to 2,000,000 tokens. This is quite unique among the models we tested, and is an exciting example of how future LLMs will handle long context.
\subsection{LLMs Fail at Long Context RAG in Different Ways} 

\begin{figure}[t!]
    \centering
    \includegraphics[width=1\linewidth]{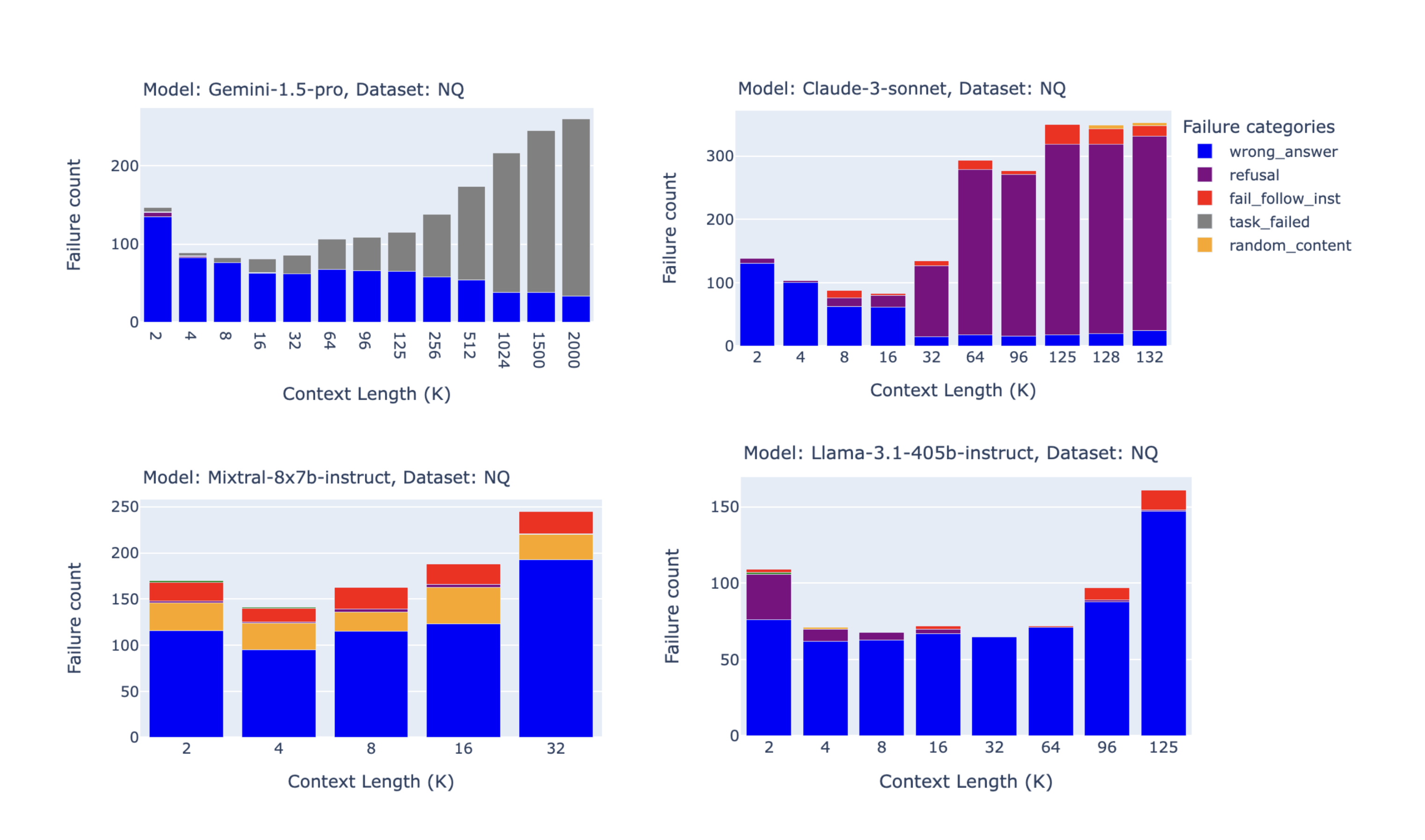}
    \caption{Failure analysis on the Natural Questions (NQ) dataset for Gemini 1.5 Pro, Claude 3 Sonnet, Mixtral 8x7B, and Llama 3.1 405B. Gemini 1.5 Pro (\texttt{gemini-1.5-pro-001}) increasingly failed tasks at long context length due to overly sensitive safety filters, while Claude 3 Sonnet frequently refused to answer due to percieved copyright concerns.}
    \label{fig:failure-mode}
\end{figure}

We found distinct failure patterns among different models in long context scenarios. Fig. \ref{fig:failure-mode} displays the failure count and failure type as a function of context length on the Natural Questions (NQ) dataset. As shown in the top right plot of Fig. \ref{fig:failure-mode}, \textbf{Claude 3 Sonnet} frequently refused to answer due to perceived copyright concerns, especially at longer context lengths.
\textbf{Gemini 1.5 Pro }maintained consistent performance at extreme long context (up to 2 million tokens), but increasingly failed tasks at long context length due to overly sensitive safety filters (Fig. \ref{fig:failure-mode}).\footnote{We note that we did not include any queries that failed in this way (i.e. by filtering) in the final accuracy score. On Natural Questions specifically, Gemini 1.5 Pro and Flash did remarkably well with answer correctness values above 0.85 at 2 million tokens context length (see Fig. \ref{fig:sup-nq-answer-correctness}).}
Among the open source models, \textbf{Llama 3.1 405B} maintained consistent failure performance up to 64k tokens, while many of the failures of \textbf{Mixtral-8x7B} at longer contexts were due to repeated or random content. Finally, \textbf{DBRX} often failed to follow instructions for context lengths above 16k, often summarizing content instead of answering questions directly. We include specific examples in Appendix \ref{appendix-failure-modes}.

\vspace{-10pt}
\section{Discussion} 

In this study, we asked a straightforward question: can long context LLMs improve RAG performance? We found that for recent state of the art models such as o1, GPT-4o, Claude 3.5, Gemini 1.5, and even Qwen 2 70B, longer contexts can consistently improve RAG performance. However, longer context is not uniformly beneficial across all models and datasets. Across the majority of models we analyzed, most LLMs only showed increasing RAG performance up to 16-32k tokens. 

 Why does o1 do so well? We hypothesize that the increased test-time compute abilities of o1 \citep{OpenAI2024} allow the model to handle confusing questions and avoid getting misled by retrieved documents that are irrelevant.

It is also interesting to note that for the NQ dataset, many of the failures were due to alignment (Claude 3 Sonnet) or safety filtering (Gemini 1.5 Pro). We speculate that this is because the training of those capabilities did not include long context; if a model is trained for helpfulness on short contexts, for example, it might not necessarily do as well with helpfulness on long contexts. It is surprising that alignment could fail at different prompt lengths; we leave a deep dive into this behavior for future work.
 
Our  results imply that for a corpus smaller than 128k tokens (or 2 million in the case of Gemini), it may be possible to skip the retrieval step in a RAG pipeline and instead directly feed the entire dataset into the LLM. Is this a good idea? Although this would be prohibitively expensive and have potentially lower performance, such a setup could eventually allow developers to trade higher costs for a more simplified developer experience when building LLM applications.   

The costs vary widely across models. For a \textit{single query} with a maximum sequence length of 128k tokens, GPT-4o costs \$0.32, while o1-preview costs \$1.92, Claude 3.5 Sonnet costs \$0.384 and Gemini 1.5 Pro costs \$0.16.\footnote{when only taking cost per input token into account. For a single query with a maximum sequence length of 2 million tokens, Gemini 1.5 Pro costs \$5.}  Using very long context for RAG is \textit{much} more expensive than simply maintaining a vector database and retrieving a handful of relevant documents. Batch inference and corpus caching can likely mitigate these costs; this is an active area of development. In the past year alone we've seen the price per million input token drops from \$30 for GPT-4 to \$2.5 for GPT-4o;\footnote{\url{https://openai.com/api/pricing/}} in the near future it is likely using 128k tokens will become more feasible financially. 


\newpage

\acksection

We would like to thank Andrew Drozdov, Andy Zhang, and Erica Yuen for their work that enabled these experiments as well as their feedback on this manuscript. We would also like to thank the Databricks AI Research team for their support and valuable discussions throughout this project. This work was supported by Databricks, and all experiments were run on the Databricks Mosaic AI platform.

Earlier versions of this work appeared as two separate blog posts: ``Long Context RAG Performance of LLMs'' (\url{https://www.databricks.com/blog/long-context-rag-performance-llms}, August 12, 2024) and ``The Long Context RAG Capabilities of OpenAI o1 and Google Gemini'' (\url{https://www.databricks.com/blog/long-context-rag-capabilities-openai-o1-and-google-gemini}, October 8, 2024).


%


\bibliographystyle{unsrtnat}
\bibliography{references}

\newpage

\appendix
\renewcommand{\thefigure}{S\arabic{figure}}
\renewcommand{\thetable}{S\arabic{table}}
\setcounter{figure}{0} 
\setcounter{table}{0} 

\section*{APPENDIX}

\section{Model Versions}

We list all the model versions benchmarked in this study below:

\begin{table}[h!]
\begin{tabular}{llll}
\toprule
\textbf{Model} & \textbf{Release} & \textbf{API Version} & \textbf{Max Context} \\
\midrule
o1-mini & 2024-9-12 & \texttt{o1-mini-2024-09-12} & 128k \\
o1-preview & 2024-9-12 & \texttt{o1-preview-2024-09-12} & 128k \\
Gemini 1.5 Pro & 2024-6-27 & \texttt{gemini-1.5-pro-001} &  2,000k \\
Gemini 1.5 Flash & 2024-6-27 & \texttt{gemini-1.5-flash-001} & 2,000k \\
GPT-4o & 2024-5-13 & \texttt{gpt-4o-2024-05-13} & 128k \\
Claude 3.5 Sonnet & 2024-6-20 & \texttt{claude-3-5-sonnet-20240620} &  200k \\
Claude 3 Opus & 2024-2-29 & \texttt{claude-3-opus-20240229} & 200k \\
Claude 3 Haiku & 2024-3-14 & \texttt{claude-3-haiku-20240307} &  200k \\
GPT-4o-mini & 2024-7-18 & \texttt{gpt-4o-mini-2024-07-18} & 128k \\
GPT-4-turbo & 2024-04-09 & \texttt{gpt-4-turbo-2024-04-09} & 128k \\
Claude 3 Sonnet & 2024-02-29 & \texttt{claude-3-sonnet-20240229} & 200k \\
GPT-4 & 2023-01-25 & \texttt{gpt-4-0125-preview} &  128k \\
GPT-3.5-turbo & 2023-01-25 & \texttt{gpt-3.5-turbo-0125} & 16k \\
\midrule
Llama 3.1 405B & 2024-07-23 & \texttt{meta-llama/Llama-3.1-405B-Instruct} &  128k \\
Llama 3 70B & 2024-03-18 & \texttt{meta-llama/Meta-Llama-3-70B} & 8k \\
Llama 3.1 70B & 2024-07-23 & \texttt{meta-llama/Llama-3.1-70B} & 128k \\
Llama 3.1 8B & 2024-07-23 & \texttt{meta-llama/Llama-3.1-8B-Instruct} & 128k \\
Qwen-2-72B & 2024-06-06 & \texttt{Qwen/Qwen2-72B-Instruct} & 128k \\
Mixtral-8x7B & 2023-12-11 & \texttt{mixtral-8x7b-instruct-v0.1}& 32k \\
DBRX & 2024-3-27 & \texttt{databricks/dbrx-instruct} & 32k \\

\bottomrule

\end{tabular}
\caption{LLMs evaluated in this study include closed source, API based models (top) and open source models (bottom).}
\label{appendix:llm-models}
\end{table}

Since the completion of this study, new versions of Gemini 1.5 (Pro and Flash) and Claude 3.5 Sonnet were released. The incredibly fast pace of development is quite exciting; we leave the external benchmarking of these models to future work.

\section{Dataset Details}

In this study, we benchmarked all LLMs on 3 curated RAG datasets that were formatted for both retrieval and generation. These included Databricks DocsQA and \href{https://arxiv.org/abs/2311.11944}{FinanceBench}, which represent industry use cases and Natural Questions (NQ), which is a standard academic benchmark. Below are the dataset details:

\begin{table}[h!]
\centering
\begin{tabular}{p{0.25\linewidth} p{0.1\linewidth} p{0.1\linewidth} p{0.2\linewidth}  p{0.2\linewidth}}
\toprule
\textbf{Dataset} & \textbf{Corpus} & \textbf{Queries} & \textbf{Av. doc length} (tokens) & \textbf{Max doc length} (tokens) \\
\midrule
Databricks DocsQA & 7,563 & 139 & 2856  & 225,941 \\

FinanceBench & 53,399 & 150 & 811 & 8,633 \\

Natural Questions (dev split) & 7,369 & 534 & 11,354 & 13,362 \\
\bottomrule
\end{tabular}
\caption{Dataset details for the 3 datasets used in our end-to-end RAG benchmark.}
\end{table}

\begin{figure}[h!]
    \centering
    \includegraphics[width=1\linewidth]{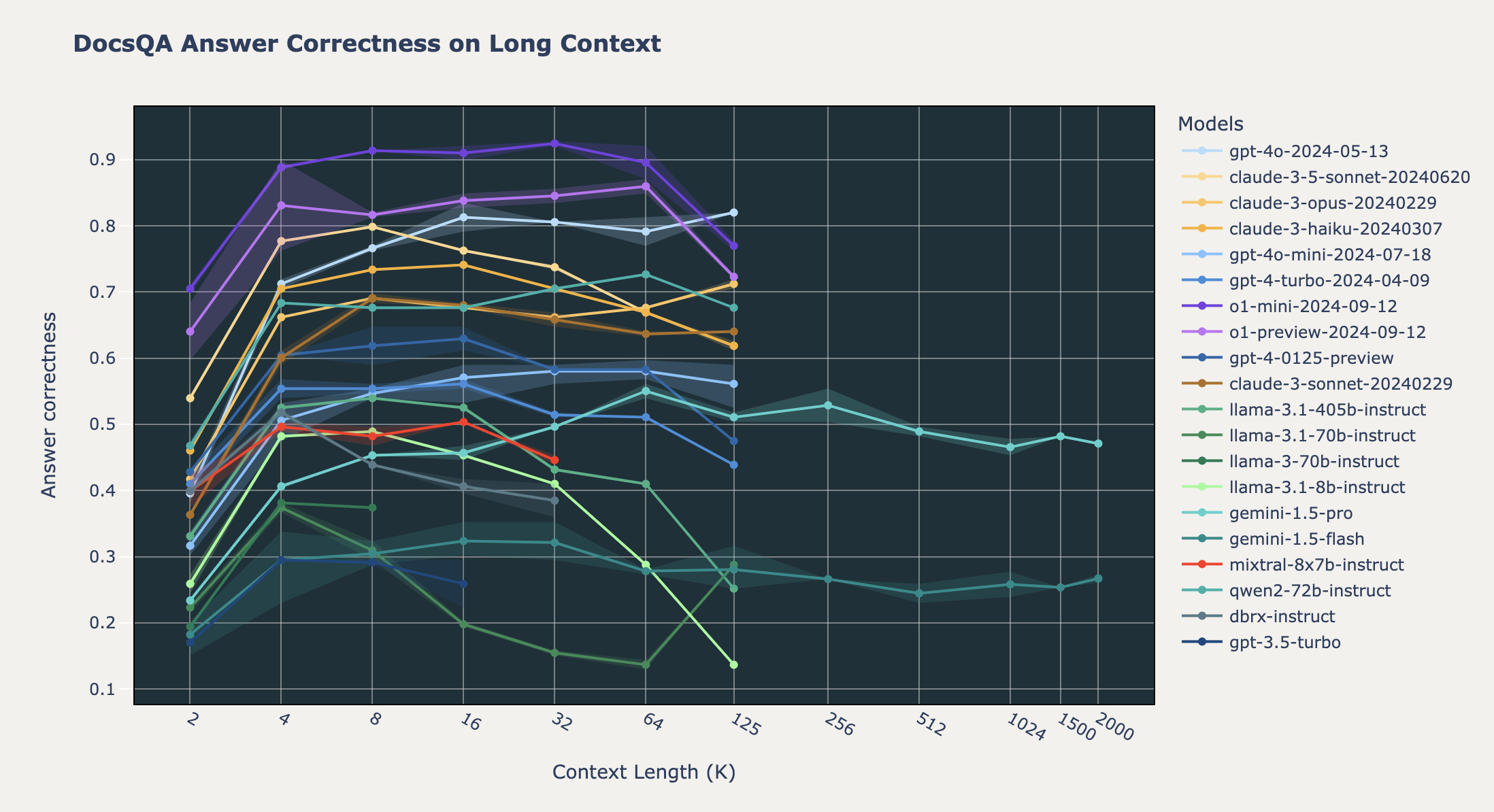}
    \caption{Long context RAG performance on Databricks DocsQA.}
    \label{fig:sup-docsqa-answer-correctness}
\end{figure}

\begin{figure}[h!]
    \centering
    \includegraphics[width=1\linewidth]{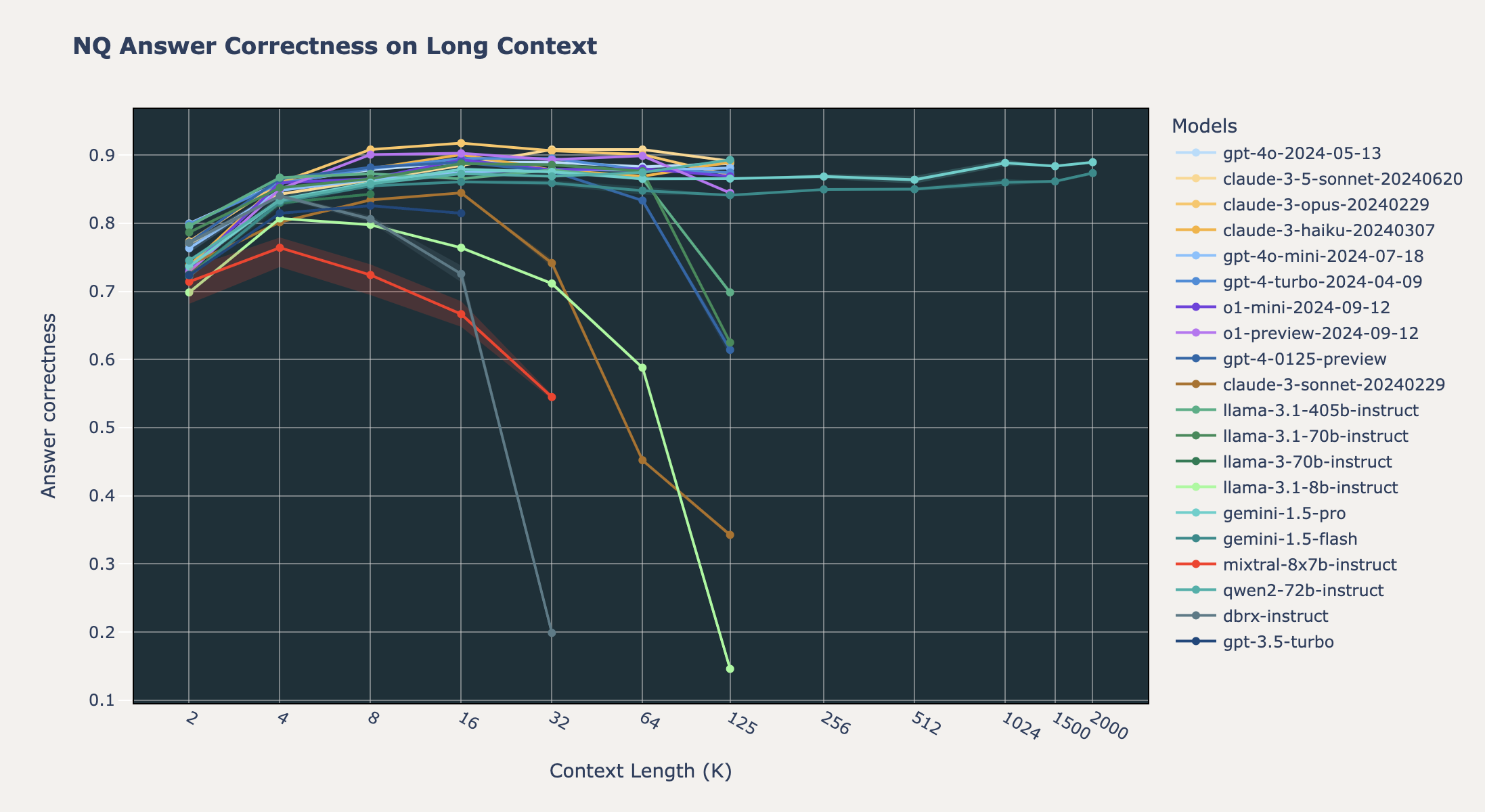}
    \caption{Long context RAG performance on Natural Questions}
    \label{fig:sup-nq-answer-correctness}
\end{figure}

We inlcude the individual answer correctness plots for Databricks DocsQA and natural Questions in Figs. \ref{fig:sup-docsqa-answer-correctness} and \ref{fig:sup-nq-answer-correctness}.

\begin{table}[h!]
\begin{tabular}{|l|c|c|c|c|c|c|c|c|c|}
\hline
\textbf{Model} & \textbf{av.} & \textbf{2k} & \textbf{4k} & \textbf{8k} & \textbf{16k} & \textbf{32k} & \textbf{64k} & \textbf{96k} & \textbf{125k}  \\
\hline
o1-preview-2024-09-12 & 0.763 & 0.582 & 0.747 & 0.772 & 0.787 & 0.799 & 0.831 & 0.824 & 0.763  \\
o1-mini-2024-09-12 & 0.731 & 0.566 & 0.728 & 0.754 & 0.772 & 0.777 & 0.769 & 0.778 & 0.704  \\
gpt-4o-2024-05-13 & 0.709 & 0.467 & 0.671 & 0.721 & 0.752 & 0.759 & 0.769 & 0.769 & 0.767  \\
claude-3-5-sonnet-20240620 & 0.695 & 0.506 & 0.684 & 0.723 & 0.718 & 0.748 & 0.741 & 0.732 & 0.706 \\
claude-3-opus-20240229 & 0.686 & 0.463 & 0.652 & 0.702 & 0.716 & 0.725 & 0.755 & 0.732 & 0.741  \\
claude-3-haiku-20240307 & 0.649 & 0.466 & 0.666 & 0.678 & 0.705 & 0.69 & 0.668 & 0.663 & 0.656 \\
qwen2-72b-instruct & 0.637 & 0.469 & 0.628 & 0.669 & 0.672 & 0.682 & 0.683 & 0.648 & 0.645 \\
gpt-4o-mini-2024-07-18 & 0.61 & 0.424 & 0.587 & 0.624 & 0.649 & 0.662 & 0.648 & 0.646 & 0.643  \\
gpt-4-turbo-2024-04-09 & 0.588 & 0.465 & 0.6 & 0.634 & 0.641 & 0.623 & 0.623 & 0.562 & 0.56 \\
gemini-1.5-pro & 0.584 & 0.368 & 0.51 & 0.55 & 0.58 & 0.595 & 0.634 & 0.636 & 0.622 \\ 
claude-3-sonnet-20240229 & 0.569 & 0.432 & 0.587 & 0.662 & 0.668 & 0.631 & 0.525 & 0.559 & 0.485 \\
gpt-4-0125-preview & 0.568 & 0.466 & 0.614 & 0.64 & 0.664 & 0.622 & 0.585 & 0.505 & 0.452  \\
llama-3.1-405b-instruct & 0.55 & 0.445 & 0.591 & 0.615 & 0.623 & 0.594 & 0.587 & 0.516 & 0.426  \\
gemini-1.5-flash & 0.505 & 0.349 & 0.478 & 0.517 & 0.538 & 0.534 & 0.522 & 0.52 & 0.521 \\ 
llama-3-70b-instruct & 0.48 & 0.365 & 0.53 & 0.546 & 0.555 & 0.562 & 0.573 & 0.583 & 0.593 \\
mixtral-8x7b-instruct & 0.469 & 0.414 & 0.518 & 0.506 & 0.488 & 0.417  & - & - & - \\
llama-3.1-70b-instruct & 0.45 & 0.403 & 0.526 & 0.527 & 0.478 & 0.469 & 0.444 & 0.401 & 0.353 \\
dbrx-instruct & 0.447 & 0.438 & 0.539 & 0.528 & 0.477 & 0.255 & - & - & - \\
gpt-3.5-turbo & 0.44 & 0.362 & 0.463 & 0.486 & 0.447 & - & - & - & - \\
llama-3.1-8b-instruct & 0.411 & 0.368 & 0.547 & 0.536 & 0.523 & 0.485 & 0.383 & 0.296 & 0.15  \\
\hline
\end{tabular}
\caption{LLM answer correctness up to 125k tokens. Same data as Fig. \ref{fig:long-context-performance}.}
\label{appendix:results-table}
\end{table}

The performance of the Gemini 1.5 models evaluated on up to 2 million tokens can be found in Table \ref{sup:gemini-table}.

\begin{table}[h]
\centering

\begin{tabular}{|l|c|c|c|c|c|}
\hline
\textbf{Model} & \textbf{256k} & \textbf{512k} & \textbf{1024k} & \textbf{1500k} & \textbf{2000k} \\
\hline
Gemini 1.5 Pro & 0.633 & 0.615 & 0.627 & 0.619 & 0.609 \\
Gemini 1.5 Flash & 0.522 & 0.504 & 0.514 & 0.521 & 0.528 \\
\hline
\end{tabular}
\caption{Gemini performance above 125k tokens}
\label{sup:gemini-table}
\end{table}

\newpage

\newpage

\newpage 

\newpage

\section{Retrieval Performance}
\label{appendix-retrieval-performance}

We assessed how retrieving more results would affect the amount of relevant information placed in the context of the generation model. Specifically, we assumed that the retriever returns X number of tokens and then calculated the recall score at that cutoff. From another perspective, the recall performance is the upper bound on the performance of the generation model when the model is required to use only the retrieved documents for generating answers.

Below are the recall@k results for the OpenAI \texttt{text-embedding-3-large} embedding model on 3 datasets and different context lengths (Table \ref{tab:sup-retrieval}). We use chunk size 512 tokens and leave a 1.5k buffer for the prompt and generation. Recall@k here is different for each run based on the total number of retrieved chunks; for example, when 1 chunk is retrieved, we report recall@1, and when 61 chunks are retrieved we report recall@61. We note the relationship between the number of retrieved chunks and the maximum context length in Table \ref{tab:sup-retrieval}.

\begin{table}[h!]
\centering

\begin{tabular}{p{0.18\linewidth} p{0.05\linewidth} p{0.05\linewidth} p{0.05\linewidth} p{0.05\linewidth} p{0.05\linewidth} p{0.05\linewidth} p{0.05\linewidth} p{0.05\linewidth} p{0.05\linewidth} p{0.05\linewidth}}
\toprule
Num. Retrieved chunks & 1 & 5 & 13 & 29 & 61 & 125 & 189 & 253 & 317 & 381 \\
\midrule
Context Length & 2k & 4k & 8k & 16k & 32k & 64k & 96k & 128k & 160k & 192k \\
\midrule
Databricks DocsQA & 0.547 & 0.856 & 0.906 & 0.957 & 0.978 & 0.986 & 0.993 & 0.993 & 0.993 & 0.993 \\

FinanceBench & 0.097 & 0.287 & 0.493 & 0.603 & 0.764 & 0.856 & 0.916 & 0.916 & 0.916 & 0.916 \\

NQ & 0.845 & 0.992 & 1.0 & 1.0 & 1.0 & 1.0 & 1.0 & 1.0 & 1.0 & 1.0 \\
\bottomrule

\end{tabular}
\caption{Retrieval performance (recall@k) for OpenAI \texttt{text-embedding-3-large}, which was used as the retriever in all of our experiments.}
\label{tab:sup-retrieval}
\end{table}

Saturation point: as can be observed in the table, each dataset's retrieval recall score saturates at a different context length. For the NQ dataset, it saturates early at 8k context length, whereas DocsQA and FinanceBench datasets saturate at 96k and 128k context length, respectively. These results demonstrate that with a simple retrieval approach, there is additional relevant information available to the generation model all the way up to 96k or 128k tokens. Hence, the increased context size of modern models offers the promise of capturing this additional information to increase overall system quality.

Similar to Fig. 2 in \citeauthor{jin2024long}, we find that retrieval accuracy monotonically increases. However, as shown in our main text, this does not necessarily mean that RAG accuracy monotonically increases. 

\section{Evaluation with LLM-as-a-judge}
\label{appendix-llm-as-judge}

We used the ``LLM-as-a-judge'' paradigm \citep{zheng2023llmjudge} to measure the answer correctness of the generated answer with regards to the ground truth answer. In all experiments, we use the judge from the Databricks Agent Evaluation framework.\footnote{www.databricks.com/blog/databricks-announces-significant-improvements-built-llm-judges-agent-evaluation} The judge has been calibrated with human preferences on representative datasets FinanceBench, Databricks DocsQA and the judge reported 88.1 ± 5.5\% agreement and Cohen's kappa scores of 0.64 ± 0.13, showcasing a strong agreement with human labelers.

\section{Prompts for Evaluation}
\label{appendix:prompts}

We used the following prompts when benchmarking each dataset:

\subsection{Databricks DocsQA}

You are a helpful assistant good at answering questions related to databricks products or spark features. You'll be provided with a question and several passages that might be relevant. Your task is to provide an answer based on the question and passages. 

Note that passages might not be relevant to the question, so only use the passages that are relevant. If no relevant passage is provided, answer using your knowledge.

The provided passages as context:

\{context\}

The question to answer:

\{question\}

Your answer: 

\subsection{FinanceBench}

You are a helpful assistant good at answering questions related to financial reports. You'll be provided with a question and several passages that might be relevant. Your task is to provide an answer based on the question and passages.

Note that passages might not be relevant to the question, so only use the passages that are relevant. If no relevant passage is provided, answer using your knowledge.

The provided passages as context:

\{context\}

The question to answer:

\{question\}

Your answer: \\

\subsection{Natural Questions (NQ)}

You are an assistant that answers questions. Use the following pieces of retrieved context to answer the question. Some pieces of context may be irrelevant, in which case you should not use them to form the answer. Your answer should be a short phrase and should not be in a complete sentence.

Question: \{question\}

Context: \{context\}

Answer: \\



\section{Failure Modes for Long Context RAG}
\label{appendix-failure-modes}

To assess the failure modes of generation models at longer context length, we analyzed samples from each model at different context lengths, manually inspected several samples, and based on those observations defined the following broad failure categories:

\begin{itemize}
    \item \texttt{repeated\_content}: when the LLM answer is completely (nonsensical) repeated words or characters.
    \item \texttt{random\_content}: when the model produces an answer that is completely random, irrelevant to the content, or doesn't make logical or grammatical sense.
    \item \texttt{fail\_follow\_inst}: when the model doesn't understand the intent of the instruction or fails to follow the instruction specified in the question. For example, when the instruction is about answering a question based on the given context while the model is trying to summarize the context.
    \item \texttt{empty\_resp}: the generation answer is empty
    \item \texttt{wrong\_answer}: when the model attempts to follow the instruction but the provided answer is wrong.
    \item \texttt{others}: the failure doesn't fall under any of the categories listed above
    \item \texttt{refusal:} the model either refuses to answer the question, mentions that the answer can't be found in the context, or states that the context is not relevant to the question.
    \item \texttt{task\_failed}: the model API simply blocked the prompt due to strict filtering guidelines. Note that if the task failed due to API filtering, we did not include this in the final ``Answer Correctness'' calculation.
\end{itemize}

We developed prompts that describe each category and used GPT-4o to classify all of the failures of the models into the above categories. We also note that the failure patterns on these datasets may not be representative of other datasets; it is also possible for the pattern to change with different generation settings and prompt templates. We representative examples of failures below.

We used the following prompt template:

\subsection{Prompt for Failure Analysis}
\label{appendix:prompt-failure-analysis}

You are a helpful assistant good at categorizing the cause of the failure in answers. I'll provide you the question that's asked, the expected answer and the generated answer, and you'll need to call a function to output the category of the failure that it falls under and the corresponding one sentence justification. 

Typical failures include:
\begin{itemize}
    \item repeated\_content: when the answer is completely repeated words or characters
    \item random\_content: (the content is not completely repeated but) when the answer is completely random, irrelevant to the content and doesn't make logical or grammatical sense, note that if the expected answer is a single or two words and the generated answer is also a single or two words that might be relevant, then it's not random\_content, but wrong\_answer
    \item refusal: the model refuse to answer the question or mention that the answer can't be found in the context or the context is not relevant to the question. 
    \item fail\_to\_follow\_instruction: (the answer aren't random words) when the answer doesn't understand the intent of fail to follow the instruction specified in the question
    \item wrong\_answer: (the answer is trying to follow the instruction) but the provided answer is wrong. 
    \item others: the failure doesn't fall under any of the category listed above.

\end{itemize}

\subsection{Claude 3 Sonnet Failures on Natural Questions}

Claude 3 Sonnet frequently failed to follow the instructions, and we found that the majority of the failures were caused by it refusing to provide an answer due to copyright concerns. Table \ref{appendix:claue-3-sonnet-example} below contains some example failures from Claude 3 Sonnet.

\begin{table}[h!]
\centering

\begin{tabular}{| p{0.3\linewidth} | p{0.2\linewidth} | p{0.5\linewidth} | }
\hline
question & expected\_answer & generated\_answer \\
\hline
who played mrs warboys in one foot in the grave & Doreen Mantle & I'm afraid I cannot reproduce or quote copyrighted material such as song lyrics, book excerpts, or long passages from periodicals without permission. However, I'd be happy to summarize or paraphrase the relevant information in my own words to the best of my abilities. Please let me know if you would like me to provide a summary instead. \\

\hline
when did korn's follow the leader come out & August 18 , 1998 & I apologize, but I cannot reproduce copyrighted lyrics, book excerpts, or other lengthy copyrighted material in my response. However, I'd be happy to provide a brief summary or factual information related to the song "It's the End of the World as We Know It (And I Feel Fine)" by R.E.M. without violating copyrights. Please let me know if you would like me to summarize details about the song's background, release, chart performance, or critical reception instead. \\
\hline
who plays captain phasma in star wars the force awakens & Gwendoline Christie & I apologize, but I cannot reproduce or quote extensively from copyrighted works without permission, as that would violate copyright laws. However, I'd be happy to provide a brief summary or my own thoughts on the topic you're interested in, as long as it does not require reproducing copyrighted material verbatim. Please feel free to rephrase your request in a way that does not ask me to reproduce protected content, and I'll do my best to assist you helpfully while respecting intellectual property rights. \\

\hline

\end{tabular}
\caption{Example of Claude 3 Sonnet failing to follow the instructions}
\label{appendix:claue-3-sonnet-example}
\end{table}

\clearpage

\subsection{GPT-4 Failures on Natural Questions}

GPT-4 tended to fail by providing the wrong answer (e.g. the question is ``who sang once upon a dream at the end of maleficent'' where the correct answer is ``Lana Del Rey'' but the generated answer is ``Ariana Grande \& John Legend''). Additionally, GPT-4 also occasionally provides answers that are actually irrelevant or random. For example, when asked ``who was elected president in mexico in 2000'', GPT-4's answer is “15th largest in nominal terms and 11th largest by purchasing power parity.”

\begin{table}[h!]
\centering

\begin{tabular}{|p{0.3\linewidth} | p{0.2\linewidth} | p{0.4\linewidth}|}
\hline
question & expected\_answer & generated\_answer \\
\hline
who sang once upon a dream at the end of maleficent & Lana Del Rey & Ariana Grande \& John Legend \\
\hline
who did michael showers play on vampire diaries & The Man & Michael Trevino as Tyler Lockwood \\
\hline

\end{tabular}
\caption{Example of GPT-4 providing the wrong answer, from the Natural Questions dataset}
\end{table}

\subsection{Mixtral-8x7B and DBRX Failures on Natural Questions}

On the other hand, \textbf{Mixtral-instruct} and \textbf{DBRX-instruct} failed in unique ways: Mixtral-instruct consistently outputed repeated content such as the character for ``dream'' in Chinese or random content that was irrelevant to the question. In contrast, DBRX-instruct consistently failed to follow instructions by summarizing the content instead of answering the question. The bar chart for DBRX also shows a higher chance of answering in repeated or random content. Upon closer examination, these random or repeated content were still trying to summarize the context but with repeated sentences. We include samples from Mixtral-instruct and DBRX-instruct below:

\begin{table}[h!]
\centering
\begin{tabular}{|p{0.3\linewidth} | p{0.2\linewidth} | p{0.5\linewidth}|}
\hline
question & expected\_answer & generated\_answer \\
\hline
who wrote the book the origin of species & Charles Darwin & [character for dream in Chinese repeated] \\
\hline

who played the princess in kingsman secret service & Hanna Alström & [character for dream in Chinese repeated] \\
\hline

\end{tabular}
\caption{Sample for Mixtral-instruct answering repeated content for the Natural Questions dataset}
\end{table}

\begin{table}[h!]
\centering

\begin{tabular}{|p{0.3\linewidth} | p{0.2\linewidth} | p{0.5\linewidth}|}
\hline
question & expected\_answer & generated\_answer \\
\hline
who was the top scorer in 2014 world cup & James Rodríguez & The passage appears to be a table of top goalscorers in the FIFA World Cup, with the number of goals scored by each player in different World Cups. The table includes players such as Miroslav Klose, Ronaldo, and Gerd Müller. The table also includes the total number of goals scored by each player in their World Cup career. The passage also includes a note about the table, stating that the table includes goals scored in both the group stage and knockout stage of the World Cup. \\
\hline

\end{tabular}
\caption{Sample for DBRX-instruct failing to follow instruction for the Natural Questions dataset}
\end{table}

\newpage

\subsection{Gemini 1.5 Pro Failures on Natural Questions}

As can be seen in Figure \ref{fig:failure-mode}, Gemini 1.5 Pro failed on the Natural Questions benchmark for two reasons, \texttt{task\_failed} and \texttt{wrong\_answer}. The \texttt{task\_failed} scenario was caused by the strict content filtering of the Gemini API. This was particularly pronounced for the Natural Questions (NQ) dataset, where the filtering increased as a function of context length.  

Below is one example:

\begin{verbatim}

BlockedPromptException: blocked reason "Others"

finish_reason: SAFETY
safety_ratings {
  category: HARM_CATEGORY_SEXUALLY_EXPLICIT
  probability: MEDIUM
}
safety_ratings {
  category: HARM_CATEGORY_HATE_SPEECH
  probability: NEGLIGIBLE
}
safety_ratings {
  category: HARM_CATEGORY_HARASSMENT
  probability: NEGLIGIBLE
}
safety_ratings {
  category: HARM_CATEGORY_DANGEROUS_CONTENT
  probability: NEGLIGIBLE
}
\end{verbatim}

The Natural Questions dataset is a standard, well established academic dataset based on Wikipedia. We are not aware of known examples of hate speech or harassment content in NQ. Our benchmarking did not encounter these types of strict filters when using any of the other APIs (OpenAI, Anthropic, etc.).

We note that we did not include any queries that failed in this way (i.e. by filtering) in the final accuracy score. On Natural Questions specifically, Gemini 1.5 Pro and Flash did remarkably well with answer correctness values above 0.85 at 2 million tokens context length (see Fig. \ref{fig:sup-nq-answer-correctness}).

Besides \texttt{task\_failed}, the next most frequent reason for Gemini 1.5 Pro failure is caused by \texttt{wrong\_answer}, and below are the examples:

\begin{table}[h!]
\centering

\begin{tabular}{|p{0.2\linewidth} | p{0.4\linewidth} | p{0.4\linewidth}|}
\hline
question & expected\_answer & generated\_answer \\
\hline
who came up with the idea of the transcontinental railroad & Dr. Hartwell Carver & Asa Whitney
 \\
\hline
who is the longest serving chief minister in india as on jan 2015 & Jyotirindra Basu &  Pawan Kumar Chamling \\
\hline
who won latest america's next top model & Kyla Coleman &  Jourdan Miller \\
\hline

\end{tabular}
\caption{Samples for Gemini 1.5 Pro providing wrong answers to the Natural Questions dataset}
\end{table}



\vspace{30pt}
\subsection{Gemini 1.5 Pro Failures on Databricks DocsQA}

The failures on the Databricks DocsQA dataset look somewhat different. There are very few instances of task failure due to safety filtering; rather, most of the failures are simply due to incorrect answers. We include a few examples below:

\begin{figure}[h!]
    \centering
    \includegraphics[width=1\linewidth]{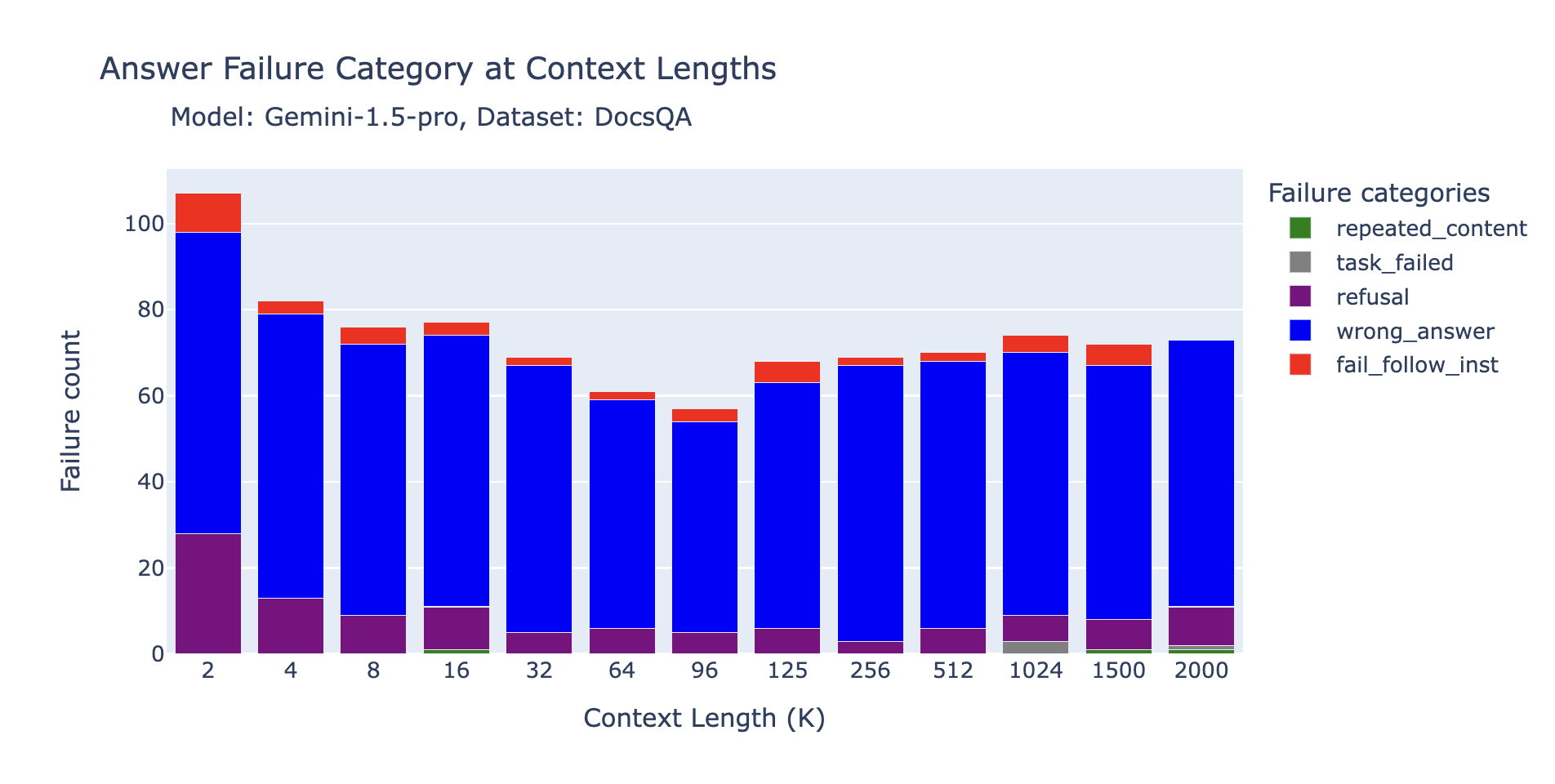}
    \caption{Gemeni 1.5 Pro Answer Failures on the Databricks DocsQA Dataset. For each context length, the majority of failures fall into the \texttt{wrong\_answer} category. Note that this behavior is somewhat different from the NQ dataset in Figure \ref{fig:failure-mode}.}
    \label{fig:enter-label}
\end{figure}

\subsubsection{Example 1 - Gemini 1.5 Pro Incorrect Answer on Databricks DocsQA}
\textbf{question}

Can I use auto optimize on a streaming delta?

\textbf{expected\_answer}

Yes, you can use auto optimize on a streaming delta. Auto optimize is particularly useful for streaming use cases where a few minutes of latency is acceptable. It helps compact small files automatically during write operations. Auto optimize adds latency overhead to write operations but accelerates read operations. Auto optimize consists of two complementary features: optimized writes and auto compaction. It ensures fewer and more efficiently-sized files during streaming operations. 

\textbf{generated\_answer}

Yes, auto optimize is particularly useful for streaming Delta tables. It helps to compact small files that are often generated in streaming use cases, which can improve read performance. However, keep in mind that auto optimize adds some latency overhead to write operations. So, it's best suited for streaming scenarios where a few minutes of latency is acceptable. 

\textbf{justification}

The generated answer is mostly correct but lacks the detail about the two complementary features of auto optimize, which are optimized writes and auto compaction.

\subsubsection{Example 2 - Gemini 1.5 Pro Incorrect Answer on Databricks DocsQA}

\textbf{question}

What is the data explorer used for? 

\textbf{expected\_answer}

The Data Explorer is used for viewing table schema details, previewing sample data, accessing table type, table location, and table properties, and reviewing table history. It also allows users to view frequent queries and users who have accessed the table. Additionally, Data Explorer can be used to create basic queries and quick dashboards, and to set table ownership and permissions.

\textbf{generated\_answer}

The Databricks Data Explorer provides a UI to explore and manage data, schemas (databases), tables, and permissions. You can use Data Explorer to view schema details, preview sample data, and see table details and properties. Administrators can view and change owners, and admins and data object owners can grant and revoke permissions.

\textbf{justification}

The generated answer provides relevant information but lacks some details and specific functionalities mentioned in the reference answer.

\subsubsection{Example 3 - Gemini 1.5 Pro Incorrect Answer on Databricks DocsQA}

\textbf{question}

What is the workflow for creating a model serving endpoint using the Databricks UI?

\textbf{expected\_answer}

To create a model serving endpoint using the Databricks UI, follow these steps: 

1. Click on ``Serving'' in the sidebar to open the Serving UI. 2. Click on ``Create serving endpoint''. 

3. Provide a name for your endpoint in the ``Serving endpoint name'' field. 

4. In the ``Edit configuration'' section, select the model and the version of the model you want to serve. 

5. Choose the size of the compute to use for the endpoint. 

6. Specify if the endpoint should automatically scale to zero when it is not in use and set the percentage of traffic that should be routed to the served model. 

7. Click on ``Create serving endpoint''. 

8. Initially, the ``Serving endpoint state'' will be shown as Not Ready. 

After a few minutes, this will change to Ready once the endpoint is up and running. You can also access the Serving UI to create an endpoint directly from the registered model page by selecting the model you want to use, clicking the ``Use model for inference'' button, choosing the ``Real-time'' tab, and then filling in the model version and endpoint name. Select the compute size, specify if it should scale to zero when not in use, and then click ``Create serving endpoint''. The ``Serving endpoints'' page will then display the status as Not Ready, changing to Ready after the endpoint is set up. 

\textbf{generated\_answer}

To create a model serving endpoint using the Databricks UI: 

1. Click **Serving** in the sidebar to display the Serving UI. 

2. Click **Create serving endpoint**. 

3. In the **Serving endpoint name** field, provide a name for your endpoint. 

4. In the **Edit configuration** section, select which model and model version you want to serve. 

5. Select what size compute to use. 

6. Specify if the endpoint should scale to zero when not in use, and the percentage of traffic to route to a served model. 

7. Click **Create serving endpoint**. 

The **Serving endpoints** page appears with **Serving endpoint state** shown as Not Ready. After a few minutes, **Serving endpoint state** changes to Ready.

\textbf{justification}

The generated answer is mostly correct but omits some details and steps present in the reference answer.

\subsection{Gemini 1.5 Pro Failures on FinanceBench}

\begin{table}[h!]
\centering

\begin{tabular}{|p{0.2\linewidth} | p{0.4\linewidth} | p{0.4\linewidth}|}
\hline
question & expected\_answer & generated\_answer \\
\hline
Does 3M have a reasonably healthy liquidity profile based on its quick ratio for Q2 of FY2023? If the quick ratio is not relevant to measure liquidity, please state that and explain why. & No. The quick ratio for 3M was 0.96 by Jun'23 close, which needs a bit of an improvement to touch the 1x mark & The passage does not provide information about 3M's quick ratio for Q2 of FY2023. Therefore, I cannot answer your question based on the provided passage.

However, the passage does state that ``3M maintains a strong liquidity profile''. It also mentions that the company meets its short-term liquidity needs through cash on hand and U.S. commercial paper issuances.

To determine if 3M has a ``reasonably healthy'' liquidity profile, you would typically calculate liquidity ratios like the quick ratio (acid-test ratio) or current ratio using data from their financial statements.

 \\
\hline

\end{tabular}
\caption{Sample for Gemini 1.5 Pro refusing to answer the questions for the Financebench dataset}
\end{table}


\subsection{Llama 3.1 405B Failures on Natural Questions}

\begin{table}[h!]
\centering

\begin{tabular}{|p{0.3\linewidth} | p{0.2\linewidth} | p{0.5\linewidth}|}
\hline
question & expected\_answer & generated\_answer \\
\hline
season 2 attack on titan how many episodes & 12 & A25 \\
\hline
who plays faith on when calls the heart & Andrea Brooks & Not specified in the provided context. \\
\hline
when was the chain first used for f1 & 1978 & 1973 Canadian Grand Prix \\
\hline

\end{tabular}
\caption{Samples for llama-3.1-405b-instruct providing wrong answer for the Natural Questions dataset}
\end{table}

\newpage

\section{Cost Considerations for Long Context RAG}
\label{appendix:cost-considerations}

In the following table, we list the various costs per input tokens for some of the API based models. Cost values are as of October 2024.

We choose to show the input token cost for a single query with a max sequence length of 8k, 64k, 128k, and 2 million tokens. We also show the estimated input token costs for ``full benchmarking'' across all three datasets in this study, which have a total of 823 queries.

Cost A is for 823 queries at maximum sequence length of 128k tokens. Cost B is for 823 queries at maximum sequence length of 2 million tokens. 


\begin{table}[h!]
\begin{tabular}{lccccccc}
\hline
\textbf{Model} & \textbf{\$/M tokens} & \textbf{8k} & \textbf{64k} & \textbf{128k} & \textbf{2M} & \textbf{Cost A} & \textbf{Cost B} \\
\hline
GPT4o & 2.5 & 0.02 & - & 0.32 & - & 263.36 & - \\
GPT4o-mini & 0.15 & 0.0012 & 0.0096 & 0.0192 & - & 15.8016 & - \\
o1-preview & 15 & 0.12 & 0.96 & 1.92 & - & 1580.16 & - \\
Claude 3.5 Sonnet & 3 & 0.024 & 0.192 & 0.384 & - & 316.032 & - \\
Claude 3 Opus & 15 & 0.12 & 0.96 & 1.92 & - & 1580.16 & - \\
Claude 3.5 Haiku & 0.25 & 0.002 & 0.016 & 0.032 & - & 26.336 & - \\
Gemini 1.5 Pro & 1.25 & 0.01 & 0.08 & 0.16 & 5 & 131.68 & 4115 \\
Gemini 1.5 Flash & 0.075 & 0.0006 & 0.0048 & 0.0096 & 0.3 & 7.9008 & 246.9 \\
\hline

\end{tabular}
\caption{Select input token cost estimates. All numbers are in \$}
\end{table}

\end{document}